\documentclass[journal,onecolumn,12pt]{IEEEtran}

\usepackage{ifpdf}
\usepackage{cite}
\usepackage[hyphens]{url}
\usepackage{graphicx}
\usepackage{booktabs}
\usepackage{threeparttable}
\usepackage[cmex10]{amsmath}
\usepackage{mathtools}%
\usepackage{algorithmic}
\usepackage{array}
\usepackage{stfloats}
\usepackage[switch]{lineno}
\usepackage{amsfonts}
\usepackage{color}
\usepackage{pifont}
\usepackage{hhline}
\usepackage{colortbl}
\usepackage{longtable}
\definecolor{Gray}{gray}{0.8}
\usepackage[]{footmisc}
\usepackage{soul}
\usepackage{wrapfig}
\usepackage{comment}

\newcommand{\bi}{\begin{itemize}}
\newcommand{\ei}{\end{itemize}}
\newcommand{\bd}{\begin{displaymath}}
\newcommand{\ed}{\end{displaymath}}
\newcommand{\be}{\begin{equation}}
\newcommand{\ee}{\end{equation}}
\newcommand{\bea}{\begin{eqnarray}}
\newcommand{\eea}{\end{eqnarray}}
\newcommand{\ba}{\begin{array}}
\newcommand{\ea}{\end{array}}
\newcommand{\bc}{\begin{center}}
\newcommand{\ec}{\end{center}}

\IEEEoverridecommandlockouts   

\title{Human movement augmentation \\ and how to make it a reality

\author{Jonathan Eden*, Mario Br\"acklein*, Jaime Ib\'a\~nez Pereda, Deren Yusuf Barsakcioglu, Giovanni Di Pino, Dario Farina, Etienne Burdet*, Carsten Mehring*}
\thanks{* denotes equal contribution. This research was supported in part by the European Commission grants H2020 NIMA (FETOPEN 899626), TRIMANUAL (MSCA 843408), as well as by the Italian Ministry of Education, University and Research grant ENABLE (FARE R16ZBLF9E3).}}

\begin{document}

\maketitle

\begin{abstract}
Augmenting the body with artificial limbs controlled concurrently to the natural limbs has long appeared in science fiction, but recent technological and neuroscientific advances have begun to make this vision possible. By allowing individuals to achieve otherwise impossible actions, this movement augmentation could revolutionize medical and industrial applications and profoundly change the way humans interact with their environment. Here, we construct a movement augmentation taxonomy through what is augmented and how it is achieved. With this framework, we analyze augmentation that extends the number of degrees-of-freedom, discuss critical features of effective augmentation such as physiological control signals, sensory feedback and learning, and propose a vision for the field.
\end{abstract}

\section{Introduction}
\begin{figure}[!t]
\centering
\includegraphics[width=\textwidth]{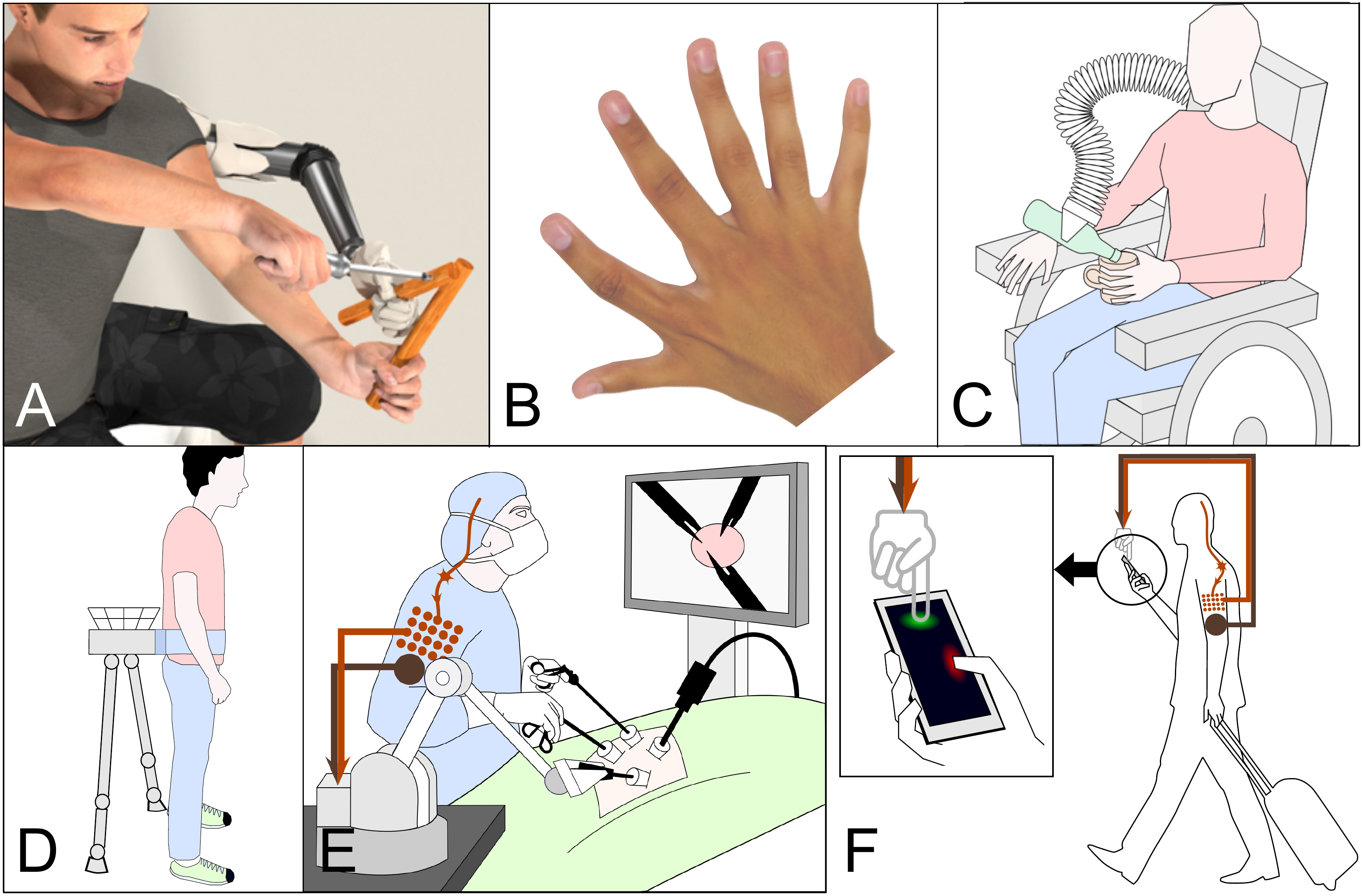}
\caption{DoF augmentation concepts and natural augmentation. A. Extra hand for assembly tasks, B. Polydactyly hand with six fingers providing superior manipulation abilities \cite{mehring2019augmented}, C. Third arm to facilitate activities of daily living in hemiplegics, D. Centaur robot for stability and walking assistance, E. Three hand surgery with hands and a neural interface, F. Augmented interaction with mobile device can free one hand to e.g. operate a map. Part A by Tobias Pistohl; Part B modified from\cite{mehring2019augmented}; Parts C, E and F by Nathanael Jarrass\'e; Part D by Camille Blondin.
}
\label{f:examples}
\end{figure}

The goal of human movement augmentation is to extend a person’s movement abilities. When this augmentation increases the number of movement degrees-of-freedom (\textit{DoF augmentation}) it can enable a person to perform tasks that are impossible to perform with their natural limbs alone. An example would be a third arm that a person can control simultaneously to the natural arms in trimanual tasks (Fig.\ref{f:examples}A). In this emerging paradigm, a user is endowed with a \textit{supernumerary effector} (SE) in the form of a wearable limb (Fig.\ref{f:examples}A), an external robot (Fig.\ref{f:examples}C,E) or a virtual effector (Fig.\ref{f:examples}F). Using technologies initially developed to compensate for motor impairments such as prosthetics and robotic exoskeletons, DoF augmentation mainly considers healthy subjects and is not bounded to natural limb appearance. Such freedom could in turn be used to develop technologies for aiding impaired individuals (Fig.\ref{f:examples}C,D).

Despite the recent growth in interest in DoF augmentation \cite{tong2021review}, the realisation of SE that can be controlled independently from the natural limbs and in coordination with them has remained elusive. A fundamental open question is whether human users can support the requirements of controlling additional DoFs without limiting natural movement. In this regard, a recent study demonstrated that subjects born with an extra finger on each hand can control the multiple extra DoFs, giving them superior manipulation abilities without any apparent movement deficits (Fig.\ref{f:examples}B, \cite{mehring2019augmented}). However, it is unclear whether subjects can learn to control artificial \textit{supernumerary DoFs} (sDoFs) that they are not born with, and whether this could enhance functional abilities. If so, where would such augmentation capabilities come from and what are their limits? How can the nervous system represent the extra limb and its relation to other limbs? These questions will impact the future of movement augmentation and determine which approaches are most suited. 

This paper analyzes the potential and constraints for different DoF augmentation strategies by considering these questions. 
Particular emphasis is placed onto the neuroscientific and technical factors that can enable user control of sDoFs, rather than the specific device design and fabrication. Section \ref{s:taxonomy} develops a taxonomy of movement augmentation, which yields the first classification of different types of augmentation. The components needed for augmentation are identified in Section \ref{s:components}. Sections \ref{s:autonomous}-\ref{s:true} review and analyze the current implementations of DoF augmentation based on the novel taxonomy, considering the potential features for each augmentation type. Sections \ref{s:feedback} and Section \ref{s:learning} then examine how the critical factors of feedback and learning affect these different augmentation types. Finally, Section \ref{s:discussion} identifies the impediments and open issues to make DoF augmentation a reality.

\section{A taxonomy of movement augmentation}   \label{s:taxonomy}
\begin{figure}[!t]
\centering
\includegraphics[width=0.97\textwidth]{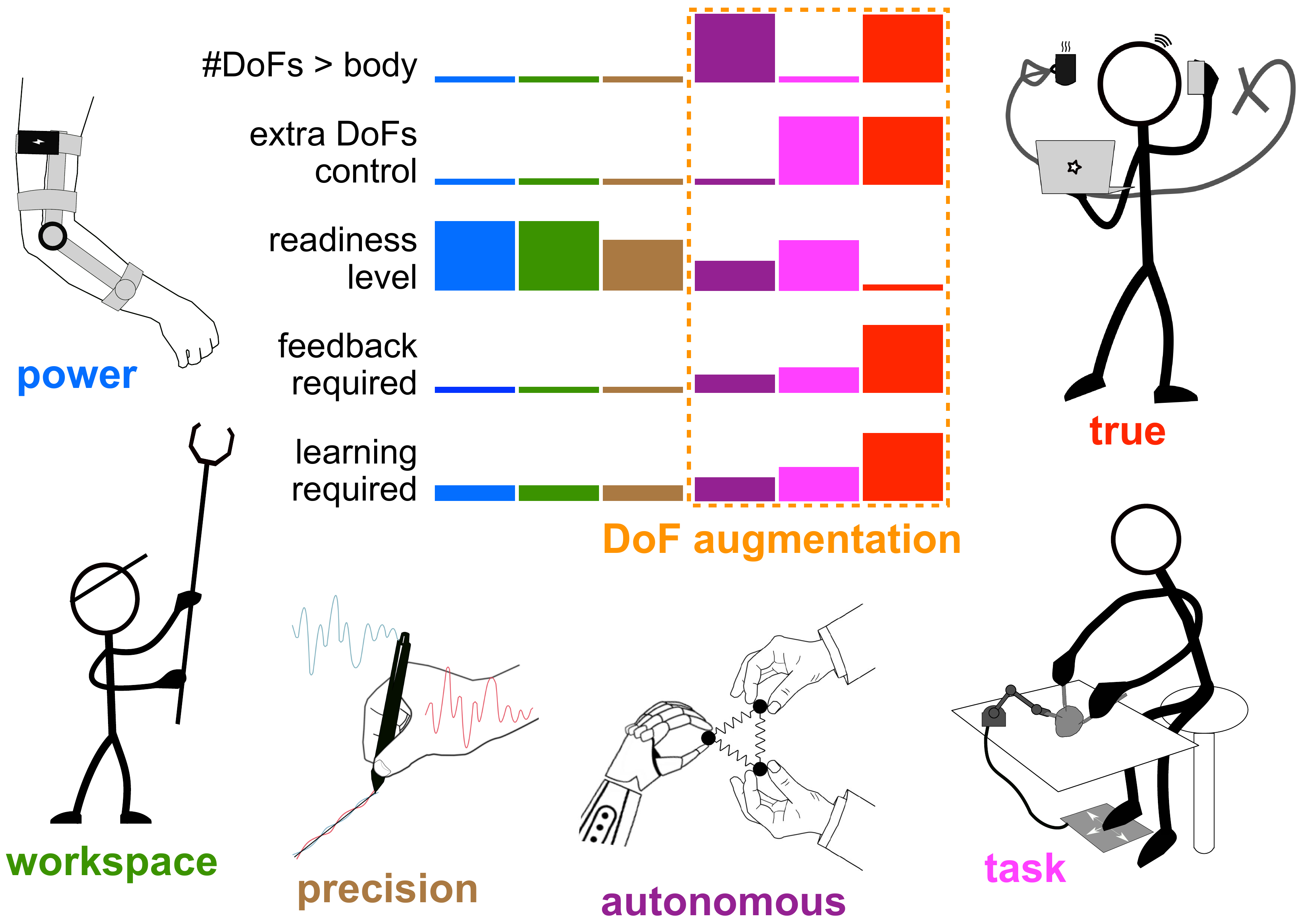}
\caption{Schematic overview of the different concepts of human movement augmentation and qualitative analysis of major characteristics. The logos for each form of augmentation (figures by Camille Blondin) illustrate example applications. The five histograms correspond to whether or not the form of augmentation grants the user more DoFs than their body and provides them with explicit control, as well as the level of how ready each device is for widespread usage, and how much feedback and learning would be required.
}
\label{f:augmentationConcepts}
\end{figure}

Movement augmentation can be classified based on the specific aspects of  motor action that are enhanced (Fig.\ref{f:augmentationConcepts}). Existing forms of augmentation include:
\begin{itemize}
\item \textit{Power augmentation} which increases the user's forces or speed. Examples are cars increasing speed as well as exoskeletons \cite{dollar2008lower} and suits \cite{zhang2017human} reducing physical load \cite{de2016exoskeletons}.
\item \textit{Workspace augmentation} which extends the spatial reach of natural limbs, with tools such as a rake or an endoscope\cite{sung2001robotic}, or through teleoperation or outer space manipulation\cite{gu1995normal}.
\item \textit{Precision augmentation} which increases the precision/accuracy during motor actions. An example includes robotic interfaces that attenuate hand tremor during eye surgery through active noise cancellation \cite{riviere2003toward}.
\end{itemize}
These 
augmentation forms improve already existing movement abilities. In contrast, DoF augmentation endows subjects with extra abilities to interact with their environment. While it has only emerged recently \cite{tong2021review}, DoF augmentation provides tremendous potential to reshape human-environment interaction as the examples of Fig.\ref{f:examples} illustrate. 

DoF augmentation ideally provides independent and coordinated control of sDoFs with the natural DoFs. Hence, a mere increase in the number of mechanical DoFs is not sufficient for DoF augmentation as the additional DoFs also need to be controlled at least to some degree independently from the natural DoFs. This can be realized in three different ways (Fig.\ref{f:augmentationConcepts}): 
\begin{itemize}
\item \textit{Autonomous augmentation} extends the number of DoFs involved in one or more tasks using autonomously controlled devices. For instance, a robot may help carry an object with its human user. 
\item \textit{Task augmentation}, in contrast, lets the user control the sDoFs. However, task augmentation only extends the number of movement DoFs involved in a task by re-purposing other body movement DoFs.  An example would be a third arm controlled by foot movements for three tool surgery.
\item \textit{True augmentation} lets the user control the sDoF by extending the body's total number of movement DoFs. An example would be a third arm controlled by neural activity that can be controlled independently from and concurrently with the natural limbs while preserving the full repertoire of natural movement abilities. 
\end{itemize}

The different ways to achieve DoF augmentation are analyzed in Fig.\ref{f:augmentationConcepts}. While all forms of DoF augmentation may provide sDoFs, only true augmentation grants the user both an increased number of movement DoFs and their direct control. The figure also illustrates the differences between these augmentation schemes in terms of their readiness to be used, their potential requirement for additional feedback devices and the learning required for their use.

The assumption of task and true augmentation is that the human user is able to voluntarily manipulate bodily signals that do not interfere with natural motion behaviors. Limbs not involved in a task, such as the foot while seated in bimanual manipulation, could  
be used to enlarge the range of possible actions, enabling task augmentation. Additionally, as the number of muscles is higher than the body’s mechanical DoFs, there is muscle redundancy that could potentially be used for task and true augmentation. Moreover, the number of neurons used for musculoskeletal control is much higher than the number of muscles, suggesting the possibility of further DoF augmentation capability. Such areas in the space of possible signals which do not correlate with differences in movements, have been coined a ``null space” \cite{nazarpour2012flexible, kaufman2014cortical, law2014rapid} in analogy to the null space of linear algebra.

\section{Features of movement augmentation} \label{s:components}
\begin{figure}[!b]
\centering
\includegraphics[width=0.3\textwidth]{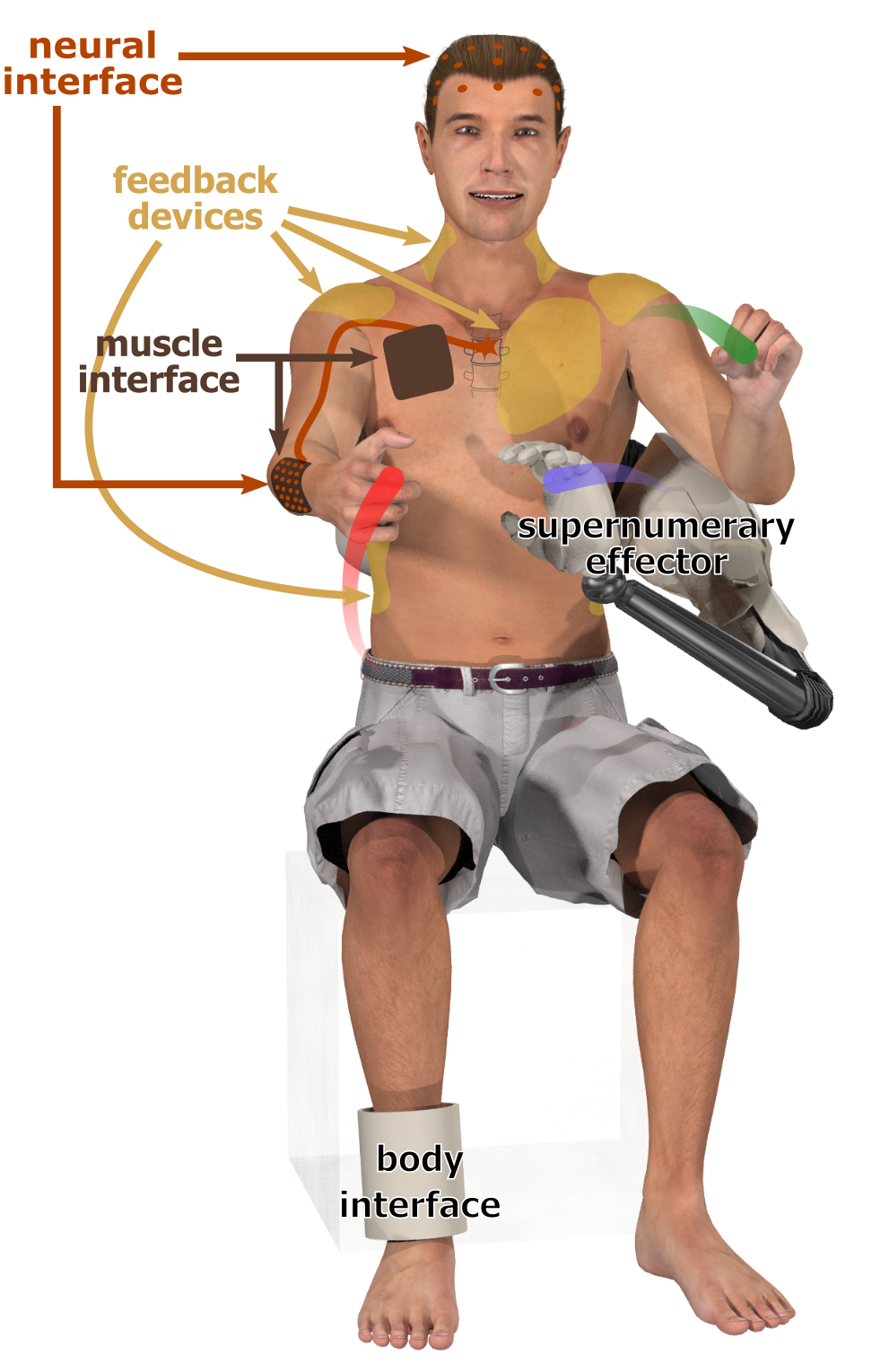} 
\caption{Interfaces for DoF augmentation 
(figure by Tobias Pistohl). An individual is augmented using a body, muscle or neural interface to control the supernumerary effector. Additional sensory feedback may be provided through feedback devices. The interfaces, supernumerary effector and feedback devices are shown at representative locations. While muscle interfaces are generally non-invasive, neural interfaces can be both invasive or non-invasive.}
\label{f:interfaces}
\end{figure}

DoF augmentation typically includes three components (Fig.\ref{f:interfaces}): The \textit{supernumerary effector} (SE) that provides the sDoFs, the \textit{command interface} that converts user intention into commands for the SE; and the \textit{feedback devices} which give the user SE status knowledge. The SE can be a robotic limb or a virtual effector. It can be wearable and move with the body (Fig.\ref{f:examples}A,D), or can be separated, e.g. a robotic arm fixed to its user's wheelchair (Fig.\ref{f:examples}C). Additionally, the technological design can be optimised to its functional task and, thus, may vary across applications. For instance a surgical device can be controlled directly by the surgeon (Fig.\ref{f:examples}E), while a mobile phone application (Fig.\ref{f:examples}F) may be controlled directly while subjects can simultaneously use their hands.

Existing SE research has mainly focused on  developing supernumerary limbs, which typically comprise robotic arms \cite{veronneau2020multifunctional, nguyen2019soft, sanchez2019four, parietti2017independent, bonilla2014robot, sasaki2017metalimbs, vatsal2018design, parietti2014supernumerary}, fingers \cite{clode2018thirdthumb, wu2014supernumerary, prattichizzo2014sixth, malvezzi2019design, hussain2017soft, cunningham2018supernumerary} and legs \cite{parietti2015design, khazoom2020supernumerary, hao2020supernumerary, treers2016design, kurek2017mantisbot}. Supernumerary arms are fixed to the user's torso \cite{veronneau2020multifunctional, nguyen2019soft, parietti2017independent}, shoulders \cite{bonilla2014robot} or elbow \cite{vatsal2018design}. Applications include aircraft fuselage assembly \cite{parietti2014supernumerary}, construction \cite{seo2016applications} and surgery \cite{abdi2016third}, while the complexity of their control has to date limited their usage. In contrast, supernumerary fingers and legs typically possess fewer sDoFs and have applications focused on aiding impaired individuals \cite{hussain2017toward} or support of gait \cite{hao2020supernumerary}. Supernumerary fingers have taken the form of either an extra thumb \cite{clode2018thirdthumb, prattichizzo2014sixth} or additional stabilising fingers \cite{wu2014supernumerary}. Virtual SEs have been applied for studying a subject's ability to use an SE \cite{abdi2016demanding, huang2020tri,bashford2018concurrent,bracklein2021towards}, or to better understand how subjects perceive augmentation through additional limbs \cite{guterstam2011illusion, sasaki2016changing, cadete2020} or fingers \cite{hoyet2016wow}.  However, to date applications such as those in Fig.\ref{f:examples}F have not been realized.

A command interface for SE control is required in many DoF augmentation applications. Three forms of interface are considered (Fig.\ref{f:interfaces}):
\begin{itemize}
\item \textit{Body interfaces} use the measured movement or force of a body segment. Body interfaces are in general non-invasive and may use limb movement, or information coming from the head such as gaze, facial expression or the tongue. 
\item \textit{Muscle interfaces} pick up muscle activity to command the SE. Non-invasive interfaces can use surface electromyography (EMG), magnetomyography (MMG), or intramuscular EMG as an invasive alternative.
\item \textit{Neural interface} extract signals from the nervous system. Non-invasive interfaces may use electroencephalography/magnetoencephalography (EEG/MEG), functional magnetic resonance imaging (fMRI), functional near-infrared spectroscopy (fNIRS) or the spiking activity of motor units. Invasive interfaces measuring signals inside the brain, spinal cord or muscles may also be used.
\end{itemize}

\section{Autonomous augmentation}       \label{s:autonomous}
In autonomous augmentation, the sDoFs are not directly controlled by the user. Instead, a software program coordinates sDoF movement\cite{bonilla2014robot}. This augmentation acts as a human-robot collaboration that can provide precise and quick movement, while minimising additional cognitive load. However, autonomous augmentation lacks knowledge of the user's desired behavior and therefore has been mainly used in specialised applications where SEs have very constrained behaviors, such as overhead assembly \cite{bonilla2014robot}, aircraft manufacturing\cite{parietti2014supernumerary}, or grasping support\cite{wu2014synnergies}. 

Autonomous augmentation's critical problems are user intent estimation and its transformation into SE action. For simple objectives such as bracing \cite{parietti2014bracing} or gait support \cite{kurek2017mantisbot}, these two problems are typically translated into control problems. In this manner, the user’s intent is assumed to be completely known and defined by the task. For example, the intent may be assumed to be to maintain a posture and then the action is that of a stabilising controller\cite{parietti2014bracing}. Such strategies provide safe and/or optimal action, but they give no flexibility for user action and therefore can only be applied to specialised activities.

For more complex behaviors, autonomous augmentation relies on predictive schemes requiring the current state, task knowledge and potentially user measurements.  Dimensional reduction or other machine learning algorithms \cite{setiawan2020grasp, wu2016implicit,wu2014synnergies} have been heavily used by supernumerary fingers for which the action of the limb can be well imagined, however, it is again limited to only work under a small set of desired actions. Recently, an alternative of considering the autonomous SE as the follower within a redundant leader-follower system was proposed \cite{khoramshahi2021intent} using an observer to estimate the user’s intent. While the method does not depend on previous data, it needs a task model meaning that it is limited to previously known actions.


Partial autonomy which splits the sDoF allowing for those with well-defined function to be autonomous while other sDoF are user controlled represents a means to benefit from autonomous augmentation while minimising its disadvantages\cite{guggenheim2020leveraging}. When and how to use it is an open research topic. In general, it is not obvious how to select useful automatic behaviors, however nature may help that choice. In humans reflexes complement slower voluntary actions where for instance long delay reflexes compensate for dynamic coupling \cite{kurtzer2008long} and contribute to controlling standing \cite{asai2009model}. The forces exerted by a wearable SL during use are not negligible and demand the user to compensate for them \cite{veronneau2020multifunctional}. When using one or more SEs in dynamic scenarios \cite{sasaki2017metalimbs, guggenheim2020laying, veronneau2020multifunctional}, the control could implement low-level automatic compensation for dynamic coupling between the natural limbs and SLs such as to ensure the body stability, so that the user could neglect them and focus on task control.

\section{Task augmentation}     \label{s:task}
In task augmentation, the human operator uses body DoFs not used in a task to control the sDoFs and coordinate them with the natural DoFs. An example of task augmentation is given by excavator control. Successful excavation requires the simultaneous movement of the platform and bucket, which is achieved through simultaneous commands from the feet and hands. 
Compared to teleoperation that transfers DoFs, task augmentation increases the task's DoFs by redirecting DoFs not normally used in performing the intended task.

Task augmentation typically uses a command input of movements \cite{abdi2015control, sasaki2017metalimbs, clode2018thirdthumb, huang2020subject} or muscle activations \cite{parietti2017independent, hussain2016emg} that do not directly interfere with the task-specific motions in which the sDoF are used. Therefore the sDoFs controlled in task augmentation are in the task’s null space (Section \ref{s:taxonomy}). Using volitional sDoF control with a suitable interface, task augmentation enables the control of complex supernumerary limbs as has been demonstrated in enhancing dexterity \cite{clode2018thirdthumb, leigh2016body}, advanced industrial settings \cite{srinivas2015multipurpose}, or through the control of a 13 DoF robotic endoscope and tool system \cite{huang2020subject, huang2020three}.

A simple task augmentation implementation is to use DoFs from direct kinematic or force recordings, for instance from a 3D camera system \cite{abdi2015control} or mechanical sliders \cite{dougherty2019evaluating}. The cognitive load associated with this coordination may place limits on the possible movement speed and accuracy \cite{abdi2016demanding, huang2020tri}. Pressure and bending sensors at the foot have commonly acted as an input source for seated tasks \cite{sasaki2017metalimbs} and a supernumerary thumb \cite{clode2018thirdthumb,kieliba2020neurocognitive}. Additional sources of input have included pressure sensors in the mouth through tongue control \cite{koike2016development}. The simultaneous use of position and force measurements demonstrates the potential of using body interfaces to control SEs with multiple DoFs \cite{sasaki2017metalimbs}.

SEs can also be commanded using an actuated exoskeleton or endpoint robotic interface. Here the interface measures position to control SEs and also provides force feedback that can facilitate control. For instance, a seated operator can use a passive foot interface placed on the ground to control four sDoF \cite{huang2020subject}. Active interfaces have been used both feet\cite{sanchez2019four}, a hand and a foot \cite{cunningham2018supernumerary} and an elbow \cite{wu2015hold}.

As task augmentation requires an interface, good performance and user comfort demand that this interface fits the user’s anatomy and neuromechanics. Therefore the interface should be adapted to the user’s characteristics such as size or movement patterns, and typical user features can inform the design \cite{huang2020subject, guggenheim2020leveraging}. Moreover, the mapping from user movement to SE command can be identified from individual's movement patterns using machine learning techniques \cite{huang2020subject}. 

Activation of muscles independent of the task can also be used, although this may result in large signal variability. For example, EMG signals from the torso have been used to control simple supernumerary arms \cite{parietti2017independent} and the Frontalis and Auricularis facial muscles have been considered to control a one DoF supernumerary finger \cite{hussain2017soft, meraz2017auricularis}. Despite muscles such as the Auricularis being independent of most motion, since they have an inherent function, we consider measurement of only their activation to be task augmentation, since that function is impaired while they are used as an input.

sDoFs control may also be achieved using signals from the null space of the limbs used in the task \cite{guggenheim2020leveraging, baldi2020exploiting}. For instance, in \cite{baldi2020exploiting} it was shown that the user’s arms could both generate natural motion and simultaneously control sDoFs. However, such systems depend on interference between the natural DoFs and sDoFs, which has not yet been experimentally evaluated. 


Finally, autonomous and task augmentation can be combined using automatic motion sequences initiated by actions in the task’s null space. Here, the user provides direct commands of intent to trigger predefined sequences for controlling the sDoFs, by using signals including voice commands \cite{vatsal2018design}, facial expressions \cite{fukuoka2019facedrive} or eye movements \cite{maimon2017towards}. Hand gestures have also been used to trigger movement of an artificial actuator attached to the user’s wrist \cite{hussain2016emg, leigh2016body}. While such systems can use more actions than is possible in typical autonomous augmentation, the control of the DoFs is still limited to activating predefined motions. 

\section{True augmentation}     \label{s:true}
In true augmentation, the body's DoF are extended independently from all natural DoFs, thereby preserving the natural movement repertoire. Autonomous augmentation used an external operator for sDOF control, while task augmentation used task irrelevant movements. True augmentation instead  uses endogenously generated physiological signals that can be modulated without interfering with natural limb control. Studies on true augmentation have only recently been conducted and a first demonstration is still lacking. A fundamental question of true augmentation is whether humans have the neural resources to control additional DoFs without limiting other functions\cite{di2014augmentation}.


Several studies have investigated different neural or muscular signals with regard to their applicability to control sDoFs concurrently to and independently from natural movement. In non-human primate studies, BCIs using firing rate of cortical neurons has been combined with simultaneous natural movement \cite{milovanovic2015simultaneous, orsborn2014closed}. Control could be established with motor cortical neurons that were not tuned but also with neurons that were tuned to the natural movements at similar performance as the BCI only task\cite{milovanovic2015simultaneous}.

In humans, 2D cursor control was achieved where one DoF was controlled by finger movements and the other by high-gamma band (70-90Hz) ECoG activity emerging from an associated motor cortical site \cite{bashford2018concurrent}. Subjects could dissociate their ECoG signals from the originally correlated finger movements and could modulate their signals largely independently from the ongoing finger movements despite the pre-experimental association between both. 

During EEG based BCI control, human subjects could simultaneously perform overt movement (natural DoF) \cite{cheung2012simultaneous, leeb2013thinking}. As EEG control signals were produced by movement imagination of limbs not involved in the performed overt movements these studies, however, follow the approach of task augmentation. In a more recent study, EEG signals evoked by grasping imagination have been used to trigger pre-programmed  grasping movements of a third robotic arm while subjects balanced a ball on a board with their two hands \cite{penaloza2018bmi}. However, the sources of the EEG control signals and the amount of simultaneity of both tasks in this study are unclear \cite{BurdetMehring2018bmi}.

\begin{figure}[!t]
\centering
\includegraphics[width=0.68\textwidth]{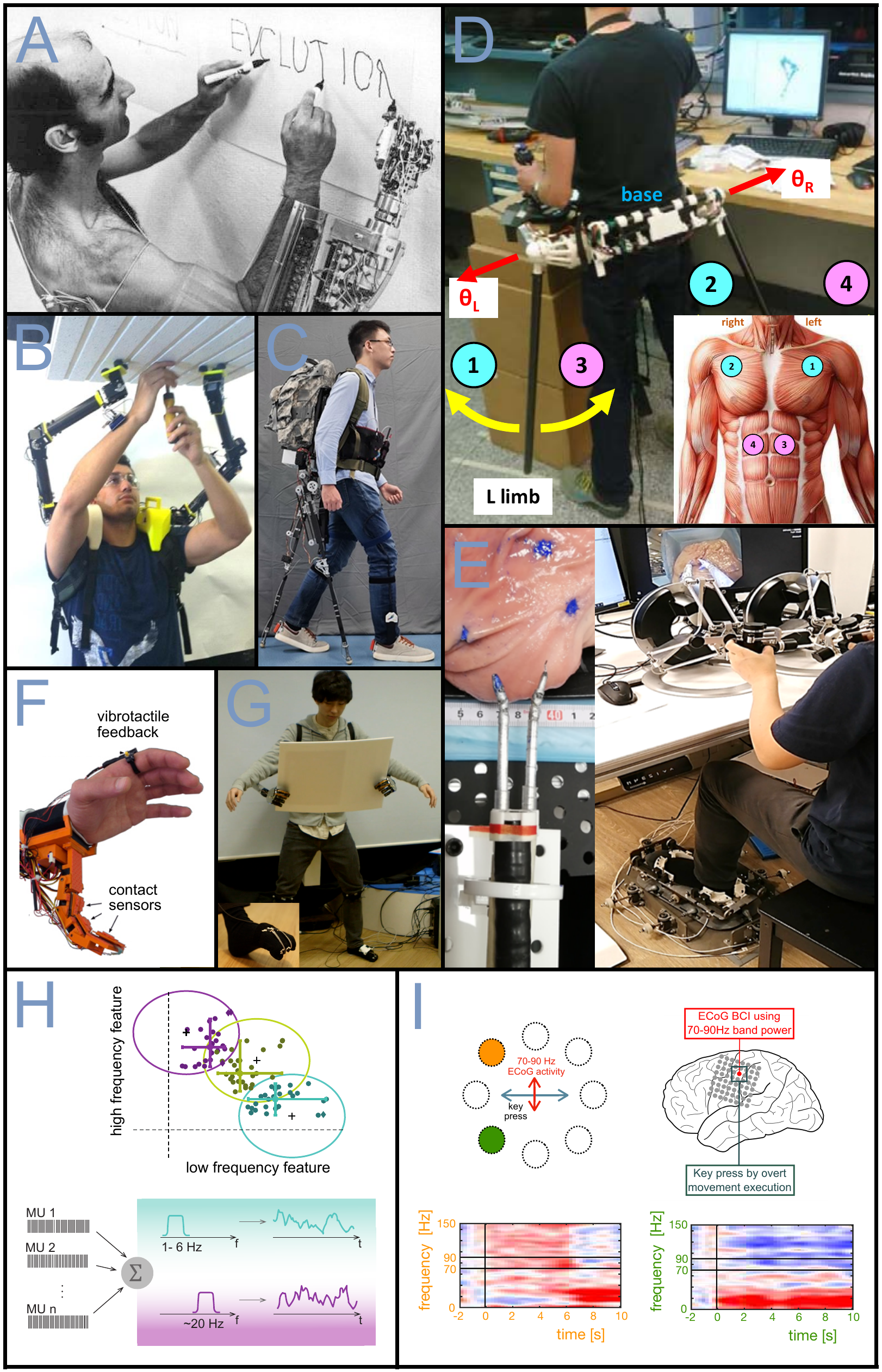}
\caption{Existing systems for DoF augmentation. An artistic vision of a third hand by Stelarc in 1980 (A, \cite{stelarc1980thirdhand}). Examples of autonomous augmentation with (B) arm to hold a part while working on it \cite{parietti2014bracing} and (C) coordinated legs to help stable walking \cite{hao2020supernumerary}. Task augmentation (D) with EMG control of supernumerary arms \cite{parietti2017independent} and (E) for three hand surgery with flexible endoscope \cite{huang2020subject}. Haptic feedback (F) for extra finger \cite{hussain2015vibrotactile} and (G) to control grasping with supernumerary arms commanded by the feet \cite{sasaki2017metalimbs}. Towards true augmentation with (H) non-invasive interface for 2D cursor control by motor neurons from the tibialis anterior muscles\cite{bracklein2021towards}. Low-frequency ($<$7Hz) and beta band (13-30Hz) activity control the horizontal and vertical cursor movements. (I) Concurrent control of high-gamma band ECoG activity and finger movement \cite{bashford2018concurrent} to control a 2D cursor. Time-resolved spectrograms show ECoG activity during movements to the upper and lower left targets. ECoG activity in the 70-90Hz band, which controls movements along the vertical axis, can be modulated independently from concurrent finger movements that control movements along the horizontal axis. Part H modified from \cite{bracklein2021towards}. Part I modified from \cite{bashford2018concurrent}.
}
\label{f:existing}
\end{figure}

For spinal motor neuron activity, obtained from high-density surface EMG \cite{barsakcioglu2020control}, human subjects could partially modulate the beta band (13-30Hz) activity without altering the force produced by the innervated muscle \cite{bracklein2021towards}. Moreover, subjects could control a 2D cursor using the low frequency ($<$7Hz) activity that is directly related to the force and the beta band activity from the motor neurons of the same muscle (see Fig.\ref{f:existing}H). Even though the beta band control remained weak, this initial study provides some support for using motor neuron populations in human movement augmentation. Moreover, recent results indicate that subjects may learn to change the motor unit recruitment order, thus, enabling 2D cursor control with three motor units from the same muscle \cite{Carmena2021}. However, subjects were instructed to perform multiple natural DoF movements to alter recruitment, which is a known way to change the recruitment order \cite{marshall2021flexible}. 

Besides neural signals, muscular activity not associated with overt movement (`muscular null space'), like for example co-contraction of muscle groups, has been used for sDoF control concurrent to isometric force generation \cite{Gurgone2021}.


Taken together, these findings provide support for the idea that sDoF control independent from natural limb movements may be realized. However in all existing studies, the overall number of controlled DoF was low, movements were simple and natural limb movement was highly restricted not reflecting its full repertoire. Moreover, in some studies different DoF were controlled by signals associated with different body parts and hence, the approach was effectively task augmentation. Thus, no study yet has demonstrated true DoF augmentation and it remains an open question whether it can be realised. 

A crucial consideration for the development of true augmentation is the choice of the physiological control signal. Besides being independent from natural limb movement and information-rich, the accessibility of the control signal is critical as true augmentation aims to provide general applications. EEG and MEG offer non-invasive brain signals, however, both recordings are prone to artefacts  and have limited resolution and bandwidth. Moreover, current MEG is not portable and EEG and MEG's usability may therefore be very limited allowing only for a small number of sDoF controlled at rather low precision and reliability. Also, fMRI and fNIRS signals may be of limited use for augmentation given their relatively low temporal resolution and in the case of fMRI their non-portability. Single units or field potentials (LFP or ECoG) may offer more information-rich and movement independent signals, yet, these signals can only be recorded invasively and may therefore not be appropriate for many applications. An emerging technique which may allow for wearable and non-invasive control of an SE is based on the spiking activities of multiple motor units obtained from surface EMG recordings \cite{farina2016principles}. Future studies will have to investigate in more detail the amount of independent control subjects can acquire over which features of spinal motor neuron firing and how this can be used for movement augmentation. Besides neural signals, muscular null space signals are an interesting alternative for movement augmentation that can also be recorded in a portable and non-invasive way and should be further investigated as a candidate signal for augmentation.


\section{Sensory feedback}  \label{s:feedback}
Representative existing systems illustrating the different categories of DoF augmentation are shown in Fig.\ref{f:existing}. Current systems rely mainly on vision to control SE performance. This may limit the user’s task focus, can require significant cognitive load, and is susceptible to occlusion. Studies on sensory feedback for prostheses have suggested that feedback plays a key role in enriching an artificial limb user's experience and control \cite{raspopovic2014restoring}. Sensory feedback for sDoFs should not substitute natural limb feedback, but extend and complement it. Furthermore, this incorporation is fundamental to achieve tentative SE embodiment, i.e. a combined internal representation together with the natural limbs which may reduce the mental effort of SE control \cite{pynn2013function}.

An often overlooked specificity of wearable SEs is that haptic feedback is intrinsically provided through the connection to the user's body as well from the motors' acoustic noise and vibration \cite{guggenheim2020leveraging}. This may be used by the brain to model and embody the SE \cite{miller2018sensing}, to acquire information on the environment \cite{franklin2008cns} or augment sensory information about the task by physical interaction with the natural limbs or collaborators \cite{takagi2017physically}. In turn, this suggests that sensory feedback is more critical for a detached robot arm used as a SE or a virtual SE (Fig.\ref{f:augmentationConcepts}).

In natural limb proprioception, the sense of presence and kinematics/dynamics of body segments are known to play a central role in movement planning and execution \cite{dadarlat2015learning}. 
Possible non-invasive feedback modalities to create similar artificial SE proprioception include vibrotactile motors \cite{alva2020wearable, noccaro2020novel},  electrotactile arrays \cite{d2013hyve, wang2019building} or direct mechanical stimulation through pressure or skin stretch \cite{wheeler2010investigation, akhtar2014passive}. 
Tactile feedback has been provided for several supernumerary hands using a direct mapping of force to haptic sensation \cite{sasaki2017metalimbs, hussain2015vibrotactile, hussain2015using}. These systems have considered one or two DoF force feedback and only a few studies have considered the effect of sensory feedback to the augmentation \cite{noccaro2020novel, hussain2015vibrotactile, wang2019building}.

A number of questions need to be investigated to develop tactile and proprioceptive feedback for DoF augmentation: Would such feedback really represent added value compared to vision? Should the feedback translate position, velocity, torque or a mixture of them? Where should such feedback be relayed? And how does the feedback affect the user’s performance and SE embodiment? The answer for each of these questions will typically differ for each form of augmentation.

Task augmentation uses the natural limbs not involved in a task to command a SE. Therefore, the user can rely on these limbs' proprioception and forward model and may not need additional sensory feedback conveyed to other body parts. To enable the association of the feedback received on the natural limbs with the interaction of the SE with the environment, haptic feedback is required e.g. through a robotic interface \cite{abdi2016third, sanchez2019four, huang2020subject}. 

In autonomous augmentation, safety of operation requires that feedback be provided if or before the SE comes in contact with the body or with the environment. However, autonomous behaviors of the SE are implemented so as to discharge the SE user from controlling it in routines tasks. Therefore, continuous feedback of the movement is not required and probably should not be provided. For example the automatic compensation for dynamic coupling with the body should not be fed back as its role is to free the user from the corresponding cognitive load and enable them to focus on the task. 

True augmentation can benefit most from a rich multimodal sensory feedback and is at the same time the most challenging class of augmentation to implement, as the motor system has no intrinsically associated sensory feedback system. True augmentation allows the use of SEs in parallel to their limbs, thus non-invasive interfaces should exploit part of the body others than limbs, such as the back or the side of the trunk, the tongue or the head. Minimally invasive interfaces could also offer a solution, such as neural interfaces implanted percutaneously \cite{oddo2016intraneural}. To compensate for neurological impairments, more invasive channels can also be considered, such as intraneural, dorsal root ganglion or epidural implants. 


\section{Learning}  \label{s:learning}
The achievable performance and skill with a SE depend on its design, the control interface and the sensory feedback it provides. 
However, the performance of augmentation will also critically depend on the learning carried out with the SE in order to improve task performance. 

Autonomous augmentation typically requires that the user comprehends the relationship between natural limb movement and autonomous behavior. As the SE’s behavior is designed to support human action, the user should normally learn to ignore the SE’s actions and fully concentrate on their relevant subtask. However, the sDoF available may result in the user selecting new strategies to perform the task, in analogy to possible behavioral changes that drivers make in adopting an automatic gearbox or ABS. Finally, for complex autonomous sequences, the human user may learn to predict the SE’s behavior, both for safety and coordination, similar to learning predictive models in movement interaction \cite{takagi2017physically}.

Task augmentation uses the natural activation of certain body segments to control the sDoFs. As humans control and coordinate their body segments from before birth, this suggests that the amount of learning to redirect control from additional natural limb to sDoFs would be limited, as long as the coordination patterns that are used are well-trained and readily available. In such cases it is mainly the mapping of the additional limbs' signals to the sDoF control that needs to be learned. The underlying learning process may correspond to the learning of modified visuo-motor coordination \cite{mazzoni2006implicit} and force fields \cite{conditt1997motor}, which can be learned quickly\cite{franklin2008cns}. Indeed, learning a simple trimanual coordination task appears to require a duration in the range of one hour \cite{abdi2016demanding, huang2020tri}. If instead new coordination patterns, which do not belong to the subject's natural movement repertoire, are used to control the primary DoFs and the sDoFs, extended practice may be required. The level of coordination between the primary DoFs and sDoFs achievable in task augmentation will also depend on the natural coordination between the associated natural limbs. For example, the coordination between one foot and hand may not be as efficient as between the two hands \cite{huang2020tri, kieliba2020neurocognitive}. Several works have investigated the learning of skilled actions such as the coordination of two extra two DoF SEs \cite{parietti2017independent} or complex manipulation with a hand equipped with a supernumerary sixth finger \cite{hussain2017soft, kieliba2020neurocognitive}. This process requires significantly more time than the simple monitor tasks such as \cite{abdi2016demanding, huang2020tri}. 

While the learning of true augmentation arguably depends on the way it is implemented, it requires that subjects can learn to modulate a control signal independently from natural movement.
Several studies have demonstrated a high degree of flexibility and adaptability of cortical neurons: Firing rates of individual motor cortical neurons  can be conditioned\cite{fetz1969operant, moritz2008direct} and as a result controlled independently of muscle activity \cite{fetz1971operant} and neighboring neurons \cite{fetz1973operantly}. New mappings from firing rates of populations of motor cortical neuron to BCI controlled cursor movement can be learned within sessions as long as the co-modulation of neurons is maintained as in natural movement\cite{Sadtler2014}. Even mappings with altered co-modulation of neurons can be learned with training spanning several days\cite{oby2019new}.
These findings demonstrate a high level of cortical neuron adaptability. They were, however, not obtained during a movement augmentation paradigm combining neural control with natural limb movements. Moreover, the reported studies were based on intracortical recordings. It remains an open question whether non-invasive recordings exhibit similar flexibility. 


A possible concern about learning to use an SE is that it may deteriorate the normal natural limb control. Could learning to control an SE overload the overall repertoire of brain function? The amount of training and the level of proficiency achieved by professional athletes or musicians would suggest that the brain is able to learn an almost unlimited number of skills. However there are reports describing how hyper-trained function can impair others, such as London taxi drivers with exceptional navigation ability at the cost of limited new spatial memory \cite{maguire2006london}. In addition to plasticity, this concern is also linked with the overall workload that can be handled by the brain, the ability to process information of which is limited \cite{townsend2011workload}. 


\section{Towards effective human movement augmentation}       \label{s:discussion}
The field of DoF augmentation has exhibited a steady increase in research activity over the last decade \cite{tong2021review}, driven by a few groups worldwide including \cite{asadagroup}. 
In this regard, a series of pioneering studies \cite{wu2014supernumerary, davenport2012design, parietti2015design, parietti2017independent, wu2014synnergies} have explored various robotic SEs and their autonomous or movement/muscle based control as well as their application to augment movement. Basic related neuroscience aspects such as natural SE \cite{mehring2019augmented}, multitasking, independence and coordination of SE and natural limbs \cite{abdi2015control, abdi2016demanding}, learning \cite{kieliba2020neurocognitive, huang2021trimanipulation}, SE embodiment \cite{guterstam2011illusion, cadete2020}, and the feasibility of true DoF augmentation \cite{bashford2018concurrent}, have also been investigated, providing foundational knowledge of user capability for augmentation. The future promises of DoF augmentation may be illustrated by amazing prototypes developed in recent years some of which are shown in Fig.\ref{f:existing}. These include an aircraft fuselage support system \cite{parietti2015design}, robotic fingers helping impaired individuals to carry out activities of daily living \cite{prattichizzo2014sixth, hussain2017soft, hussain2015vibrotactile}, gait assistance devices \cite{hao2020supernumerary}, the skilled wearable two arm system of \cite{sasaki2017metalimbs}, and a soft endoscope with two surgical tools commanded by foot and hands \cite{huang2020three}. In this section we analyze the limitations of current systems and propose several areas where we think future work should focus to make DoF augmentation a reality.

\subsection{Autonomous augmentation}
Autonomous augmentation represents a form of human-robot interaction which has been little investigated. Current systems have been restricted to simple and well constrained applications. To understand what kinds of SE behaviors could be implemented, we consider the interaction framework of \cite{jarrasse2012framework}. A strategy that has been used to implement autonomous augmentation consists of dividing the task in independent subtasks for the human and the SE. For instance, a surgeon is in charge of the whole operation but for automatic knot tying that the robot takes over. Another relatively simple autonomous augmentation strategy consists of the ``assistance behavior'' of \cite{jarrasse2012framework}, where the SE control is strictly subordinate to NL motion. For instance, to manipulate a large object with three hands, the SE would coordinate its movements to maintain shape or a constant force with respect to the natural hands. However, more complex interaction control strategies could be used where the SE is considered as an autonomous agent. For instance a third arm may be used for robot-assisted physical rehabilitation according to an egalitarian control scheme \cite{li2019differential}. Rich interactive behaviors of an autonomous SE with the NLs may be implemented based on a differential game theory framework \cite{jarrasse2012framework, li2019differential}.

Given that most applications of movement augmentation will require contact between a human and an artificial mechanism, the safety of that interaction must be considered before future applications are possible.  While this is an issue to be considered in all forms of augmentation, its impact is likely greatest in autonomous augmentation for which the user lacks direct control of the SE. Also, computer-controlled reflex mechanisms should be developed to prevent the SE  from harming the user or a nearby person.

\subsection{Task augmentation}
Task augmentation relies on the notion that users possess the capacity to simultaneously and independently control multiple limbs or muscles. Volitional modulation of control signals needs to be carried out together with ongoing movement. While the body can simultaneously control multiple DoFs, for instance coordinated motion of the hands and feet, movement augmentation may increase task complexity and require multi-tasking. The brain may have limits on the total number of DoFs it can control as well as on the complexity and number of sub-tasks it can carry out simultaneously. These limits will need to be experimentally investigated to determine the possible performance. 

Some applications will require augmented sensory feedback, for which basic questions need to be investigated. This includes the questions of what, where and how exactly it should be? Ideally since sensory feedback closes the loop from the user's action to the SE's reaction, its placement and modality should parallel the actuation and usage of the SE. The level of knowledge given by inherent feedback also needs further investigation to identify when and in which cases additional feedback modules are best suited.

Both multi-tasking and the incorporation of sensory feedback may improve with increased user experience. If learning to control sDoFs corresponds to skill learning, what can be achieved will also depend on the amount of training time required, e.g. months or years of training may be required for good performance as in sports. However, such extended training periods may not be available or desirable for all applications. Systematic research is therefore required to develop and optimise learning paradigms for acquisition of skilled SE control. 

\subsection{True augmentation}
The major current limiting factor in true augmentation lies with determining where the resources for controlling the sDoFs could come from. Two basic questions need to be addressed in this regard: i) Do users have sufficient independent physiological control signals? and ii) Can such signals be reliably and robustly sourced? Given the high degree of redundancy present from the neural to kinematic levels of the musculoskeletal system we believe that in principle the nervous system is able to learn to generate signals that can be modulated independent from movement. However, this is currently unknown, and even if the brain can generate such a signal, its reliability and dimensionality may limit the functionality of true augmentation.

Motor unit activities may offer a non-invasive and portable solution to provide multi-dimensional signals to control an SE, however it remains to be shown they can be modulated independently from movement. Moreover, current decomposition algorithms extracting spiking activity from sEMG recordings in isometric condition need to be extended to movements. Muscular null space signals are another non-invasive and portable alternative signal type for augmentation with initial results encouraging further research. Both motor unit and the muscle signals for augmentation can only be generated in the presence of some contraction of the corresponding muscles. This further challenge would need to be overcome for many applications of these signals for augmentation.

While many of these issues may be overcome in the future, the same issues on multitasking, sensory feedback and learning as described in the previous section for task augmentation need to be considered for true augmentation.
To maximise the performance of augmented movement, future research should thus determine and document the limits of human capability and user specific optimized algorithms to translate physiological control signals into augmented movement. Yet, there will likely be hard limits to the complexity and number of tasks the brain can perform simultaneously, thus restricting the functional capabilities that can be achieved with true and task augmentation.

\subsection{Experiments and Applications}
Common to all forms of augmentation is a need for greater understanding of the underlying computational and neural mechanisms. Progress in these basic aspects of movement augmentation will require extensive experiments with human subjects to analyse: i) the coordination and learning behaviors, and ii) subjective report on augmentation systems (e.g. on the aspects  of comfort, sense of agency, utility, and on how their use modify actions experience). A systematic analysis of these factors may require a framework of different possible scenarios as described in the Supplementary Materials.

Limitations in the precision and control capabilities of the SE's movements and in the accuracy of user's control signal, may limit the usability of augmentation in certain applications such as in surgery requiring precise control. These system and human limitations should be considered to design new protocols specifically for multimanual operation.

Finally, virtual SEs e.g. on computers or mobile devices are yet unexplored and may offer new and interesting possibilities for future studies as applications in augmented and virtual reality become more common.

\bibliographystyle{IEEEtran}
\bibliography{IEEEabrv,references}

\section*{Supplementary Materials Application Scenarios}

DoF augmentation has been applied to a range of different applications within medical, industrial and commercial settings. However, it is not yet clear which tasks are most suited for DoF augmentation due to a lack of understanding of the capability of human user's to exploit it. To study the effect of the coordination between the SEs and natural limbs, we propose to identify tasks based upon i) whether the limbs have to work together to achieve the task, and ii) whether there is any physical/virtual linkage between the DoFs. Table \ref{tbl:taxonomy} summarises the coordination types in the case of manipulation with three limbs with examples of typical tasks. This involves typical three-handed tasks such as holding an elastic membrane, opening a door while holding a box with both hands and manoeuvring both the camera and tools in surgery. From this taxonomy, tasks and the relative performance of augmentation may be able to be evaluated.

{\small
\begin{center}
\center
\begin{longtable}{|l|l|l|l|}
\hline
\multicolumn{1}{|c|}{} &
\multicolumn{1}{|c|}{all coupled} & \multicolumn{1}{|c|}{two coupled} & \multicolumn{1}{|c|}{all uncoupled} \\
\hline
\multicolumn{1}{|l|}{all dependent} & 
\begin{minipage}[t]{0.25\columnwidth}
manipulating an elastic membrane\\
\end{minipage} &
\begin{minipage}[t]{0.25\columnwidth}
camera for endoscopy \\
\end{minipage} &
\begin{minipage}[t]{0.27\columnwidth}
triangulation (e.g. of laser beams)
\end{minipage}\\
\hline 
\multicolumn{1}{|l|}{two dependent} & 
\begin{minipage}[t]{0.25\columnwidth}
drawing on a table with the balance of the table maintained by two arms
\end{minipage} &
\begin{minipage}[t]{0.25\columnwidth}
holding a box with two arms while opening a door with the other limb
\end{minipage} &
\begin{minipage}[t]{0.27\columnwidth}
tying shoelaces while holding an object
\end{minipage}\\
\hline 
\multicolumn{1}{|l|}{all independent} & 
\begin{minipage}[t]{0.25\columnwidth}
N/A
\end{minipage} &
\begin{minipage}[t]{0.25\columnwidth}
N/A
\end{minipage} &
\begin{minipage}[t]{0.27\columnwidth}
multi-object pick and place
\end{minipage}\\
\hline 
\caption{Coordination types of one SE and two natural limbs with representative tasks.}
\label{tbl:taxonomy}
\end{longtable}
\end{center}
}

Within this context, existing research has shown that without the presence of physical constraints (all uncoupled)  superior performance is possible using three hands in place of two \cite{abdi2016demanding, dougherty2019evaluating}. This is observed both when all hands are independent \cite{abdi2016demanding}, and when the sDoFs need to be coordinated with the natural DoF for operating a camera for position control (all dependent) \cite{dougherty2019evaluating}, as would be the case for surgery with a camera. In \cite{abdi2016demanding} subjects also felt no additional mental effort for control, and in fact expressed a preference for using the sDoF in the task. When considering tasks which can only be performed with three hands, it was observed that subjects felt little change in mental effort irrespective of the condition when going from a bimanual reaching task to a trimanual task \cite{huang2020tri}. However, such coupling does result in reduced performance and when continuously tracking additionally leads to increased cognitive load\cite{huang2020tri}. Finally, when comparing subject performance to that of dyads, the dyad has to date outperformed the single user controlling an SE who had higher mental effort in all tested cases \cite{noccaro2021human, huang2021trimanipulation}. However, the relative difference between results appears to reduce in coupled motion (all coupled all dependent) \cite{noccaro2021human}, and might disappear when haptic feedback and some learning is provided \cite{huang2021trimanipulation}, although users still reported increased mental effort.

\end{document}